\documentclass[a4paper, 10pt, conference]{ieeeconf}      

\IEEEoverridecommandlockouts                              

\usepackage{graphics} 
\usepackage{epsfig} 
\usepackage{times} 
\usepackage{amsmath} 
\usepackage{amssymb}  
\usepackage{color}

\usepackage{standalone}
\usepackage{booktabs}
\usepackage{hyperref}
\usepackage[table]{xcolor}
\usepackage[english]{babel}
\usepackage{layouts}
\usepackage{tikz,bm,color}
\usepackage{tikzscale}
\usepackage{float}
\usepackage{multirow}
\usepackage{subcaption}
\usepackage{rotating}
\usepackage{pgfplots}
\usetikzlibrary{positioning}
\usetikzlibrary{shapes,arrows}
\usetikzlibrary{decorations.pathreplacing,angles,quotes}
\usetikzlibrary{calc}
\usetikzlibrary{external} 
\DeclareMathOperator*{\argmax}{argmax} 

\colorlet{lightgray}{white!100}
\colorlet{box_gray}{gray!20}
\colorlet{gray20}{gray!20}
\colorlet{gray40}{gray!40}
\colorlet{gray60}{gray!60}
\colorlet{gray80}{gray!80}
\colorlet{gray100}{gray!100}

\title{\LARGE \bf
Towards Corner Case Detection for Autonomous Driving
}

\author{Jan-Aike Bolte$^{\ast}$, Andreas B\"ar$^{\ast}$, Daniel Lipinski$^{\circ}$ and Tim Fingscheidt$^{\ast}$
	\thanks{$^{\ast}$Jan-Aike Bolte, Andreas B\"ar and Tim Fingscheidt are with the Institute for Communications Technology, Technische Universit{\"a}t Braunschweig, Schleinitzstr. 22, 38106 Braunschweig, Germany {\tt\small \{j.bolte, a.baer, t.fingscheidt\}@tu-bs.de}}
	\thanks{$^{\circ}$Daniel Lipinski is with Volkswagen Group Research - Automated Driving, Berliner Ring 2, 38440 Wolfsburg, Germany	
		{\tt\small daniel.lipinski@volkswagen.de}}
}

\begin{document}

\maketitle
\thispagestyle{empty}
\pagestyle{empty}

\begin{abstract}
The progress in autonomous driving is also due to the increased availability of vast amounts of training data for the underlying machine learning approaches. Machine learning systems are generally known to lack robustness, e.g., if the training data did rarely or not at all cover critical situations. The challenging task of corner case detection in video, which is also somehow related to unusual event or anomaly detection, aims at detecting these unusual situations, which could become critical, and to communicate this to the autonomous driving system (online use case). Such a system, however, could be also used in offline mode to screen vast amounts of data and select only the relevant situations for storing and (re)training machine learning algorithms. So far, the approaches for corner case detection have been limited to videos recorded from a fixed camera, mostly for security surveillance. In this paper, we provide a formal definition of a corner case and propose a system framework for both the online and the offline use case that can handle video signals from front cameras of a naturally moving vehicle and can output a corner case score. 
\end{abstract}
\section{INTRODUCTION}
The recent developments in machine learning also led to significant advancements in current autonomous driving systems. These systems more and more rely  on deep learning techniques that use huge datasets for training, e.g., learning from this data how to behave in certain situations. The use of \textit{black-box} deep learning systems poses a risk, which became apparent through the latest real-world accidents with autonomous cars being involved \cite{NTSB2018,NTSB2018a}.
Such accidents may occur, if the training data did rarely or not at all cover certain situations~\cite{Tian2018}, a typical case, when issues due to machine learning can be expected~\cite{Koopman2016}. The goal of a corner case detection system is to detect unusual situations\footnote{In defining a "situation", we follow the definition by Ulbrich et al.\ \cite{Ulbrich2015}, "a situation is the entirety of circumstances, which are to be considered for the selection of an appropriate behaviour pattern at a particular point of time."} either in this training data, or, in a second step, online in an autonomous vehicle.
The video signal from a monocular vehicle camera represents a part of the situation. For humans it is possible to distinguish normal from unusual events, even when they only have available the video from a mono vehicle camera. Therefore, it should also be possible to design a corner case detection system that can evaluate situations based on video data only.\\ 
Corner case detection brings advantages both in offline as well as in online systems. In an online approach the corner case detection system can be employed as a redundant warning function accompanying an autonomous driving system, where it provides information about how unusual the current situation is (see Fig. 1 (a)). In an offline approach, the corner case detection system can be used to parse through vast amounts of collected video data and returns only a user-defined amount of unusual data. This can be used as a data selection procedure for large-scale data recordings, where it is undesirable to store too much redundant or irrelevant data. These selected corner cases can then be used for a more focused training of autonomous driving systems, tackling the problem of underrepresented critical training data, e.g., by oversampling the corner cases \cite{Chawla2002}.\\
\begin{figure}[tp!]
	\begin{subfigure}[b]{0.9\columnwidth}
		\includegraphics{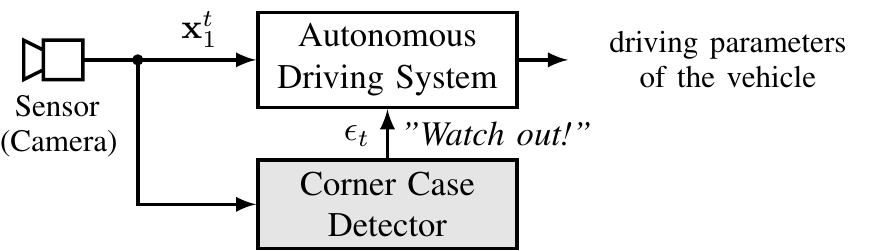}
		\vspace{-10pt}
		\caption{Online corner case detector}
		\label{fig:1}
	\end{subfigure}
	\begin{subfigure}[b]{0.9\columnwidth}
		\vspace{10pt}
		\includegraphics{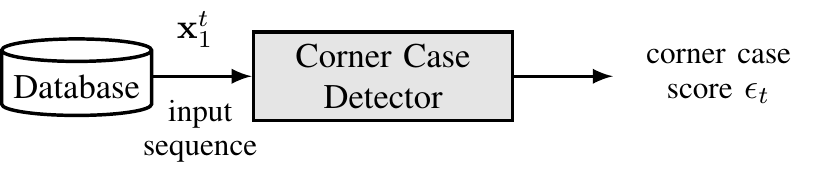}
		\vspace{-10pt}
		\caption{Offline corner case detector}
		\label{fig:2}
	\end{subfigure}
	\caption{High-level block diagrams visualizing the (a) online approach and (b) offline approach of a corner case detection system, with $x_1^t$ being the input (video) sequence until discrete time increment $t$, and $\epsilon_t$ being a corner case score at time $t$. }
\end{figure}
A major challenge in the development of a corner case detection system is that the detection of corner cases is an ill-defined problem, since there is no universally accepted definition of what a corner case actually is. Corner case detection is closely related to anomaly detection~\cite{Chandola2009} and novelty detection~\cite{Pimentel2014}, and these two disciplines are already very close to each other. Anomaly detection typically refers to the detection of samples during inference that do not conform with an expected normal behavior. To figure out what is abnormal, a world model can be trained on normal data so that any deviation from that learned behavior will be marked as an anomaly. Novelty detection is similar in the sense that a world model is trained on normal data. Any data that deviates from the already seen data is marked as a novelty. These novelties can be either anomalies or just samples that were not included in the training data. If enough diverse training data is available to train a world model that represents normality sufficiently well, any novelty will be an anomaly.  If not, any anomaly is a novelty, but not any novelty is necessarily an anomaly.\\
The core idea of our work is that the online corner case detection provides self-awareness and a criticality measure for a perception module by an advantageous combination of a semantic segmentation with a video-based prediction error, under the hypothesis that \textit{non-predictive relevant objects in vicinity} to the car's expected future trajectory is a corner case. Our contribution with this work is threefold, as we propose a formal definition of what a corner case in the context of perception in autonomous driving might be, secondly, we point out domain-specific challenges regarding corner case or anomaly detection in videos from car-mounted cameras that record scenes with a highly dynamic content and challenging ego motion. Thirdly, we propose a technical framework that will allow to develop corner case detection systems.\\
The paper is structured as follows. In Section II we present related work in the field of anomaly detection, image prediction, and semantic segmentation. In Section III we propose a definition of a corner case. In Section IV we introduce the corner case detection system concept along with employed datasets and measures. Finally we will present the results of some first experiments in Section V, before we provide conclusions in Section VI.
\section{RELATED WORK}
The corner case detection system that is proposed in this work has a modular structure, consisting of subsystems, which can be (partly) adopted from other research fields. As mentioned in Section~1, we employ an image prediction method in combination with a semantic segmentation to tackle the problem of corner case detection, being closely related to anomaly detection.
\subsection{Anomaly Detection}
Literature offers a wide variety of approaches for the detection of unusual events. Most of them fall under the terms of anomaly detection or novelty detection~\cite{Hodge2004,Chandola2009, Pimentel2014}. Driven by the rapid development in the field of deep learning, many recent methods follow the \textit{reconstruction-based} approach by \textit{unsupervised} training of neural networks, exploiting the important advantage that vast mounts of unlabeled data can be used in training~\cite{Thompson2002,Singh2004,Mahadevan2010,Hasan2016, Carrera2017,Chong2017,Utkin2017,Munawar2017, Liu2018}. The unsupervised training is typically applied to datasets that contain a large number of normal samples and a negligible small amount of abnormal samples. We assume that we are given a sequence  $\mathbf{x}_1^T = \left(\mathbf{x}_{t=1},\mathbf{x}_{t=2},...,\mathbf{x}_{t=T} \right) $, where $\mathbf{x}_t$ denotes the frame at the discrete frame index $t$, and $T$ is the length of the sequence. In this case the reconstruction-based method learns a model of normality $\mathcal{M}_{\text{normal}}$, given the training dataset $\mathcal{X}_{\text{train}}$, consisting of several sequences. The trained model is used to assign a novelty or abnormality score to the test dataset  $\mathcal{X}_{\text{test}}$. 
Various forms of autoencoder (AE) networks can be used to train such a model. \\
There are also some special approaches to anomaly detection in the context of autonomous driving or unmanned vehicles. Lin et al.\ \cite{Lin2010} used the Mahalanobis distance to measure the distance of mutliple sensor data vectors, thereby identifying the unusual events. However, their approach was not image-based and thus is not suitable for our work. Another approach is to employ particle filtering and maximum likelihood methods for anomaly detection \cite{Lampiri2017}, but again this approach is not suitable for image-based anomaly detection.\\
As a more general approach to anomaly detection, Munawar et al.\ \cite{Munawar2017} trained a spatio-temporal anomaly detection system for surveillance of industrial robots using a deep convolutional neural network (DCNN). They use a biologically plausible system for anomaly detection according to Egner et al.\ \cite{Egner2010}, where an anomaly is determined on the basis of expectation and surprise. \textit{Accordingly, an unusual video frame can be identified by its deviation from the predicted frame.} Liu et al.\ \cite{Liu2018} used a future frame prediction with spatial and motion constraints based on generative adversarial networks (GANs). \\
Encouraged by these results, we will choose a simple reconstruction-based approach but will employ a predictively trained model of normality. Different to all image-based approaches mentioned above that were evaluated on datasets with a stationary camera, the ego motion of the vehicle makes the datasets in autonomous driving research much more challenging. We therefore will now consider various image prediction approaches for automotive datasets.
\subsection{Image Prediction}
The task of predicting future frames in videos with parametric models that were trained in an unsupervised manner was rarely approached before the work of Ranzato et al.\ \cite{Ranzato2014}. Inspired by language modeling they introduced a baseline approach for extrapolating future frames that used features learned from video signals in an unsupervised fashion. It seemed to have been the first model that was able to generate realistic predictions of video sequences. Srivastava et al.\ \cite{Srivastava2015} introduced a long short-term memory autoencoder (LSTM-AE) that performed predictions on image patches, or so-called percepts, that are extracted by a preceding convolutional neural network (CNN). They showed that the predictive training even improved results in a classification task for action recognition. Mathieu et al.\ \cite{Mathieu2015} used a multi-scale architecture and an objective function that incorporated an adversarial loss and the differences in image gradients. P\u{a}tr\u{a}ucean et al.\ \cite{Patraucean2015} used a convolutional LSTM-AE. They designed a spatio-temporal video autoencoder to emulate the human visual short-term memory in a basic form. The first real model for long-term prediction was introduced by Lu et al.\ \cite{Lu2017}. They designed a novel objective function and an autoencoder structure with LSTMs.
However, again, all predictive approaches mentioned before have in common that they only evaluate their models on datasets with stationary cameras. In the context of automotive image prediction the well-known prediction network \texttt{PredNet}~\cite{Lotter2016} can capture key aspects of both movement of the ego vehicle and movement of the  objects in the visual sequences.\\
In our experiments, we will evaluate \texttt{PredNet} and a predictive autoencoder that is based on the network proposed in \cite{Hasan2016}, since the authors already used it for anomaly detection with a reconstruction-based approach. We will adopt the network and will propose an adversarially trained predictive approach for corner case detection.
\subsection{Semantic Segmentation}
The aim of semantic segmentation is the semantic labeling of each pixel of an input image.
Current state-of-the-art architectures for semantic segmentation rely on the concept of fully convolutional networks (FCNs), introduced by~\cite{Long2015}. Here, a classifier, pre-trained on the ImageNet database~\cite{Russakovsky2015}, is modified for semantic segmentation. Typical classifiers in state-of-the-art models for semantic segmentation~\cite{Zhao2016a,Chen2017,Wu2016,Chen2018a,RotaBulo2018} are residual networks (\texttt{ResNets})~\cite{He2016}. The appropriate amount of context is crucial for semantic segmentation.
One common way of dealing with this problem is the use of dilated convolution \cite{Yu2016} to enlarge the receptive field in deeper layers~\cite{Zhao2016a,Chen2017,Wu2016,Chen2018a,RotaBulo2018}.
Further multi-scale context is addressed by some state-of-the-art architectures~\cite{Zhao2016a,Chen2017,Chen2018a,Chen2018,RotaBulo2018} via combining the feature extractor with a pyramid pooling module. To restore the original resolution, all models are designed in a encoder-decoder fashion.
One simple approach is to use bilinear interpolation of the network prediction as the decoder part~\cite{Zhao2016a,Chen2017,Chen2018,RotaBulo2018}, while other approaches use more complex operations such as transposed convolution~\cite{Wu2016,Romera2018}, or the additional use of low-level features through skip connections~\cite{Chen2018a,Ronneberger2015}. \\ 
Recently, along with a major interest in practical implementations, efficient semantic segmentation architectures with regard to computation and memory cost have been introduced. To address the memory problem,~\cite{RotaBulo2018} proposes in-place activated batch normalization (\texttt{InPlace-ABN}), a memory-efficient approach in the training process through combining the leaky rectified linear unit (leaky ReLU) with batch normalization~\cite{Ioffe2015}. In~\cite{Chen2018a,Sandler2018} depthwise separable convolutions are used to reduce the number of parameters and therefore computation cost and memory usage, while~\cite{Romera2018} proposes factorized convolution in combination with residual connections to obtain a similar effect. \\
We base our own segmentation network on the \texttt{DeepLabv3}~\cite{Chen2017} with some improvements, since it is one of the best performing networks on the Cityscapes dataset.
\section{CORNER CASE DEFINITION}
A consequence of the so-far missing universally accepted definition of a corner case is that there is also no explicit metric existing. Motivated by~\cite{Munawar2017}, we will use a predictive approach for corner case detection.
The idea is that if a novel or abnormal or critical suddenly occurring situation is technically predictable, it will not pose a major problem to an autonomous driving system, which will then naturally take care of adequate actions. Therefore, for us this would be a \textit{don't care} situation, although some might view it as a corner case. So our focus is on \textit{technically unpredictable} situations.
However, it is important to note that not each unpredictable situation in the field of autonomous driving is necessarily a corner case. 
An aircraft that suddenly enters the camera image in the sky may not be predictable, but luckily in most cases it will be irrelevant for the driving task. As opposed to that, pedestrians, cyclists and other moving objects on the ground are highly relevant classes. Beyond that, even a pedestrian acting in some highly unpredictable manner may be irrelevant for the driving task if it happens at a sufficiently large distance from the vehicle or its future trajectory.\\
We therefore propose that a corner case is given if there is a \emph{relevant} object (class) in \textit{relevant} location that a modern autonomous driving system \textit{cannot predict}. Therefore, the relevance of a corner case results from the three aspects noted in Figure 2. The relevance of objects and locations, as well as aspects of prediction, will be discussed in the next section. 

\begin{figure}[tp!]
	\centerline{A corner case is given, if there is a}
	\centering
	$\underbrace{\text{\emph{non-predictable}}}_{1}
	\underbrace{\vphantom{p}\text{\emph{relevant object/class}}}_{2} \text{in} 
	\underbrace{\vphantom{p}\text{\emph{relevant location}.}}_{3}$
 	\\ \par  \noindent
 	\caption{Definition of a corner case for autonomous driving research and systems.}
 	\label{fig:coca}
\end{figure}
\noindent
\section{TOWARDS CORNER CASE DETECTION: FRAMEWORK AND MEASURES}
As a working hypothesis we assume that it is possible to detect corner cases with a camera-based system, because, as already mentioned, it is also possible for a human to distinguish normal from unusual events, even when only the video from a mono vehicle camera is available. We limit this work to the detection of non-predictive situations in the context of autonomous driving, meaning a camera in movement. 
Following our  definition of a corner case we need a system that combines the three important aspects (Fig.~2): (1) First, we need an image prediction that gives us the prediction errors for each new frame. (2) Second, we need a semantic segmentation of the input frame that allows us to classify and localize the objects in the scene, with moving objects being considered as \textit{relevant} (see Table 1), and  (3) third, we need a detection system that processes the information from both image prediction and semantic segmentation by information fusion, comprising a check, whether the non-predictable (e.g., jumping) relevant class (e.g., pedestrian) is in a \textit{relevant location} (will cross trajectory). The following subsections describe each part of the corner case detector that is illustrated in Figure 3, along with the datasets we employ for training, and related measures.
\subsection{Semantic Segmentation}
Semantic segmentation of images aims at finding a transformation that partitions the input image into semantically related parts, by assigning each pixel to a specific class. As motivated in Section II, we adopt a segmentation network based on the \texttt{DeepLabv3}~\cite{Chen2017} with some improvements from \texttt{WideResNet38}~\cite{Wu2016}, which is pre-trained on the ImageNet corpus. 
To be more specific, we replace \texttt{ResNet50}~\cite{He2016} inside \texttt{DeepLabv3} by \texttt{WideResNet38}.
This is similiar to the approach in~\cite{RotaBulo2018}, with the difference, that we don't incorporate the proposed \texttt{InPlace-ABN}. Further, we do a few common modifications in semantic segmentation to \texttt{WideResNet38}~\cite{Chen2017,Wu2016}: First, we remove the classification layer of \texttt{WideResNet38} and connect the remaining network to the segmentation head of \texttt{DeepLabv3}. Second, to control the output stride (ratio between input resolution and output resolution) we decrease the stride of several convolutions from two to one in a bottom-up fashion and increase the dilation rate instead. Third, in contrast to~\cite{Wu2016}, we do not incorporate dropout in our segmentation framework as we observed lower performance results otherwise. \\
The input image  $\mathbf{x}_t \in \mathbb{G}^{H \times W \times C}$ with image pixel $x_t(i) \in \mathbb{G}$, where $\mathbb{G}$ is the set of gray values, $H$ and $W$ are the image height and width in pixels and $C=3$ is the number of color channels from set $\mathcal{C}=\{1,2,3\}$, is  fed into a fully-convolutional neural network. It maps the input to output scores $\mathbf{P}_t \in \mathbb{I}^{H \times W \times  	\vert \mathcal{S} \vert  }$, where $\mathcal{S}$ denotes the set of classes with cardinality $|\mathcal{S}|=19$ and $\mathbb{I}= [0,1]$.
For each pixel position $i \in \mathcal{I}$, the third dimension of the output scores provides a posterior probability (score) $P_t(i,s)$ for each class $s \in \mathcal{S}$. Here, $\mathcal{I}$ is the set of pixel indices in the image, and $|\mathcal{I}| = H \cdot W$ the number of pixels. Taking the argmax over the output scores we obtain the $(H \times W)\text{-dimensional}$ mask $ \mathbf{m}_t = \argmax_{s\in\mathcal{S}} \mathbf{P}_t $, which gives us a pixel-wise classification $m_t(i)\in\mathcal{S}$ of the frame for time $t$. \\
\begin{table}[tp!]
	\centering
	\caption{The classes $s \in \mathcal{S}$ used in semantic segmentation. Within this work, we limit ourselves to mobile objects $s \in \mathcal{S}_{\mathrm{rel}} = \{12,13,...,19 \}$, which determine the \emph{relevant classes} for corner case detection; they are printed in \textbf{bold}:} 
	\resizebox{\columnwidth}{!}{%
		\begin{tabular}{clclcl}\toprule
			No.&Class&No.&Class&No.&Class\\
			\midrule
			1 & Road 		& 8 	& Traffic sign 	&  \textbf{15} & \textbf{Truck} \\
			2 & Sidewalk 	& 9 	& Vegetation 	&\textbf{16} & \textbf{Bus} \\
			3 & Building 	& 10 	& Terrain 		& \textbf{17} & \textbf{Train} \\
			4 & Wall 		& 11 	& Sky 			&  \textbf{18} &  \textbf{Motorcycle} \\
			5 & Fence 		&  \textbf{12} 	&  \textbf{Person }		&  \textbf{19} & \textbf{Bicycle} \\
			6 & Pole 		&  \textbf{13} 	&  \textbf{Rider} 		&  &  \\
			7 & Traffic light & \textbf{14} 	&  \textbf{Car} 			&  &  \\
			\bottomrule
	\end{tabular}}
	\label{tab:labels}	
\end{table} 
As a metric, the mean intersection over union (mIoU) is employed \cite{Everingham2015}, which measures the accuracy of the segmentation mask and is defined as the mean of the frame-wise 
\begin{equation}
\text{IoU}_t = \frac{\text{TP}_t}{\text{TP}_t+\text{FP}_t+\text{FN}_t} \, ,
\end{equation} 
where $\text{TP}_t\text{, FP}_t\text{, and FN}_t$ are the numbers of true positive, false positive, and false negative pixels, respectively, in frame~$t$.\\
Modern neural networks have been shown to be overconfident in their classifications \cite{Gao2017}. This can pose a problem on system components used for semantic segmentation or object detection, since even unknown objects will be classified as one of the known classes. In a future corner case detection system this will be solved by a single-frame anomaly detection.
\subsection{Image Prediction}
\begin{figure}[tb!]
	\centering
	\includegraphics{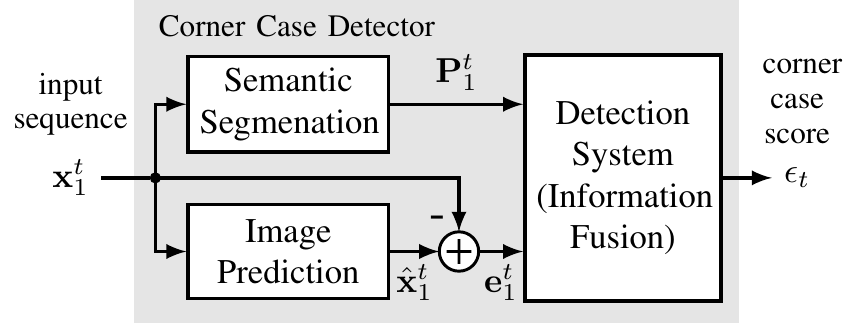}
	\caption{High-level block diagram of the corner case detector.}
	\label{fig:short}
\end{figure}
As already mentioned, image prediction is an essential part of the corner case detection system. Modern autonomous driving systems already predict trajectories of other traffic participants. To identify corner cases in video streams, it is essential to understand the underlying states and dynamics within the given situations. This high-level abstraction can be learned by predictive models.
For the image prediction approach we train a model that receives $n$ consecutive frames $\mathbf{x}_{t-n}^{t-1}= (\mathbf{x}_{t-n}, \mathbf{x}_{t-n+1},...,\mathbf{x}_{t-1})$ to compute a prediction $\hat{\mathbf{x}}_t$ of the current frame. As a metric for the corner case, we now calculate an error 
\begin{equation}
\mathbf{e}_t = \hat{\mathbf{x}}_{t} - \mathbf{x}_t,
\end{equation}
between the predicted image and the actual image $\mathbf{x}_t$ (subtraction symbol in Figure~2), with elements $e_t(i), i \in \mathcal{I}$.  
Following Mathieu et al.\ \cite{Mathieu2015}, we use the following metrics to evaluate the prediction models.
The mean-squared error (MSE) distance is then given by
\begin{align}
D_\text{MSE}(\hat{\mathbf{x}}_{t}, \mathbf{x}_{t}) &=\frac{1}{|\mathcal{I}|} \sum_{i \in \mathcal{I}}e_t^2(i).
\end{align}
Additionally we employ the peak signal-to-noise ratio (PSNR), which is defined by
\begin{equation}
\text{PSNR}(\hat{\mathbf{x}}_t, \mathbf{x}_t) = 10\log_{10} \frac{\rm{max}^2_{\hat{\mathbf{x}}}}{D_\text{MSE}(\hat{\mathbf{x}}_t, \mathbf{x}_t)},
\end{equation}  
where $\rm{max}^2_{\hat{\mathbf{x}}}$ shall denote the squared maximum possible value of the color channel image intensities, e.g., $255^2$ for 8-bit image formats with $\mathbb{G}=\{0,1,...,255\}$. Both metrics can also be applied to color images, where the metric is evaluated for each color channel separately, and then averaged.
As a third metric, we use the structural similarity index measure (SSIM), which is a metric for perceived image quality being introduced by Wang et al.\ \cite{Wang2004}. The SSIM measures the perceptual difference between the original and the predicted image and is, unlike the other two metrics, based on visible structures in the image.
\subsection{Detection System}
In the detection system, information from both of the two previous processing steps is combined. As its output the system generates a corner case score $\epsilon_t \in \left[0,1 \right] $ for each input frame $\mathbf{x}_t$, exploiting also past frames $\mathbf{x}_{t-n}^{t-1}$. Along with $\epsilon_t$, in principle also a localization of the corner case in the image can be performed. If we recall the definition of the corner case in Figure 1, we remember that we consider a corner case consisting of the logical AND combination of three aspects. The semantic segmentation provides the class and location information and the image prediction provides the \textit{predictability}. 
To identify a \textit{relevant location}, in a real system, typically one would use a perception approach based on a 
light detection and ranging (LIDAR) sensor to assign depth information to the image pixels \cite{Zhang2015,Wei2018}. On the basis of that, a time to collision on a pixel basis can be estimated \cite{Bosnak2017}.
For the purpose of this work, however, instead we adopt a rather simple approach for the reason of conciseness of presentation and evaluation. We simply assume that objects being further above the bottomline of the image are more distant to the ego vehicle. \\
The error map (2) from the image prediction gives us a value of non-predictability for each pixel. Since we are for now only interested in the moving classes, we simply set the error of those pixels that do not belong to one of these moving classes to zero, given the following formula:
\begin{equation}
\label{eq:comb_seg_pred}
{e}_{t,\mathrm{rel}}(i) = \begin{cases}
{e}_t(i), & m_t(i)\in\mathcal{S}_{\mathrm{rel}} \\
0, & m_t(i)\notin\mathcal{S}_{\mathrm{rel}}
\end{cases}
\end{equation}
where $\mathcal{S}_{\mathrm{rel}}$ denotes the set of all relevant classes for the corner case detection. In our case, we use the  eight classes that are printed in bold in Table I. The squared errors of the relevant classes ${e}_{t,\mathrm{rel}}^2(i)$ are then weighted depending on their distance from the bottom of the image and summed up resulting in an error score 
\begin{equation}
\epsilon_t^{\prime} = \sum_{i \in \mathcal{I}} {e}_{t,\mathrm{rel}}^2(i) \cdot (1-\frac{h_i}{H-1}),
\end{equation}
with $h_i \in \{ 0,1,...,H\! - \!1 \}$ being the row index (bottom-up) of pixel $i$. Thereby, our simple definition of a relevant location weights the bottom row squared errors by a one, and the top row squared errors of the relevant classes by a zero. Finally, the corner case score is obtained by normalizing the error score $\epsilon_t^{\prime}$ to a value range from 0 to 1 using
\begin{equation}
\epsilon_t = \frac{\epsilon_t^{\prime}-\min\limits_{\tau \in \mathcal{T}} \epsilon_{\tau}^{\prime}}{\max\limits_{\tau \in \mathcal{T}} \epsilon_{\tau}^{\prime}-\min\limits_{\tau \in \mathcal{T}} \epsilon_{\tau}^{\prime}} ,
\end{equation}
where $\mathcal{T}$ denotes a set of time instants. For the online approach it may be $\mathcal{T}=\{1,2,...,t\}$, or all time instants of the validation data. For the offline approach $\mathcal{T}$ may contain all time instants of the video material currently being analyzed. 
If a \textit{localization} of the  corner case in the frame is needed, $\epsilon_t^{\prime}$ is obtained by (6) with summation over small patches $i \in \mathcal{I}_p \subset \mathcal{I}$ of the window. We therefore subdivide the image into patches of the same size, e.g., $32\! \times \! 32$ pixels. The patch size can be adjusted by the user, depending on how exactly he desires to localize the corner cases. The error scores of each of the patches can then also be normalized to a range between 0 and 1. \\
Finally, the obtained corner case score is subject to thresholding. An appropriate threshold value $0<\theta_\epsilon<1$ has to be identified on validation data to tune the desired behavior of the detector regarding the false acceptance rate and the false rejection rate. In an offline system, $\theta_\epsilon$ can be chosen by the user in order to control the amount of detected corner cases in the data. \\
\section{DATASET, TRAINING, AND QUALITATIVE EVALUATION OF THE CORNER CASE DETECTION SYSTEM FRAMEWORK}
\subsection{Dataset}
We train both the segmentation and image prediction network on the Cityscapes dataset~\cite{Cordts2016} that contains a diverse set of street scene images recorded in 50 different cities, being a widely used benchmark for semantic segmentation not only for autonomous driving research. The dataset is labeled with 19 classes that are used during training and inference (see Table 1). They denote the set $\mathcal{S}$.
The Cityscapes dataset \cite{Cordts2016} offers a benchmark suite that serves as a baseline for future improvements in image segmentation. For the purpose of this conceptual paper, the image prediction is trained on the three demo videos provided by the dataset.
\subsection{Semantic Segmentation}
We mainly adopted the training protocol from Chen et al.\ \cite{Chen2017}.
For optimization, we used the stochastic gradient descent with momentum $\beta=0.9$ and a learning rate with polynomial decay:
\begin{equation}
\label{eq:lr_schedule}
\eta\left( k\right) = \eta_{\mathrm{start}} \cdot \left( 1-\dfrac{k}{k_{\mathrm{max}}}\right)^{\gamma}
\end{equation}
where $\eta\left(k\right)$ is the current learning rate at iteration $k$, $\eta_{\mathrm{start}}$ is the initial learning rate, $k_{\mathrm{max}}$ the maximum number of iterations, and exponent $\gamma=0.9$.
For data augmentation we perform random resizing in the range $\left[ 0.5, 2.0\right]$, left-right flipping, and cropping of the input image with a crop size of 700x700.
With this configuration and our reimplementations we were able to fit a batch of size $b_1=4$ and $b_2=2$ using an output stride of $o_1=16$ and $o_2=8$, respectively, on an \texttt{Nvidia Geforce GTX 1080 Ti}.

The training itself is organized in a two-stage fashion.
In the first stage, we set the output stride to $o_1=16$ and train the network parameters, including the batch statistics, for 90,000 iterations with a batch of size $b_1=4$ and an initial learning rate $\eta_{\mathrm{start}}=0.001$.
In the second stage, we set the output stride to $o_2=8$, freeze the batch statistics in the corresponding layers, and fine tune for an additional 120,000 iterations with a reduced batch of size $b_2=2$ and a reduced initial learning rate $\eta_{\mathrm{start}}=0.0005$.

We evaluate our segmentation results by using the fine-tuned stage-two model and computing the mean intersection over union (mIoU, see (1)) between the network prediction and the ground truth with an output stride of $o=8$.
We use multiple parallel fine-tuned models, which are fed by input images with three different scales $q \in \mathcal{Q}=\lbrace0.75, 1.0, 1.25 \rbrace$, resulting in score maps for each class $s \in \mathcal{S}$ and scale $q \in \mathcal{Q}$.
These score maps are resized to the original input size and are fused by summation for each class $s \in \mathcal{S}$ separately across all scales $q \in \mathcal{Q}$. 

Our network was solely trained on the finely annotated training images from the Cityscapes dataset. It achieves a competitive mIoU of 78.4\% on the Cityscapes evaluation set, being quite close to the best approaches known today.
 
\subsection{Image Prediction}
For the image prediction task we tested two models to find out which one works best for the difficult conditions in automotive applications. The first model is the well-known \texttt{PredNet} \cite{Lotter2016} and the second is an autoencoder network that is based on the network proposed by Hasan et al.\ \cite{Hasan2016}.\\ 
To gain a quick first insight into the employment of image prediction the three demo videos provided by the Cityscapes dataset were used as training data for both prediction networks. 
\begin{figure}[tp!]
	\centering
	\includegraphics{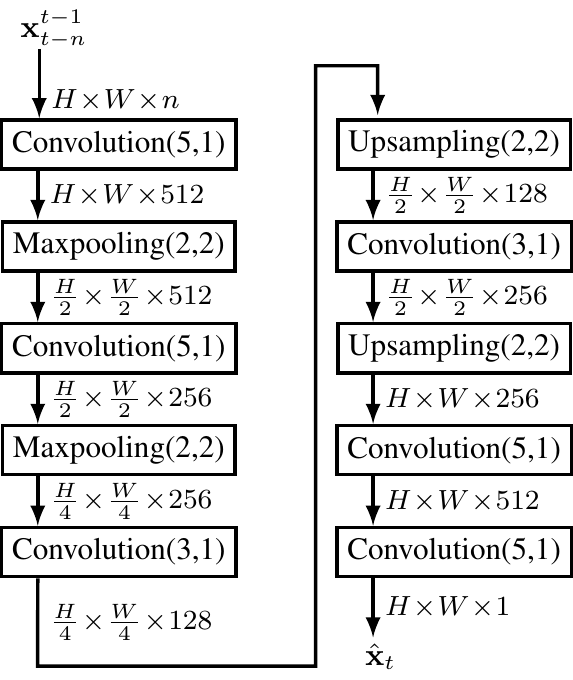}
	\caption{Architecture of the \textbf{image prediction} by an autoencoder. A convolution, maxpooling, or upsampling block with parameters ($K,S$) uses a filter size of $K \times K$ and a stride of $S$. The annotations at the edges indicate the dimensions of the transferred tensors.}
	\label{fig:short}
\end{figure}
Both networks were trained on the three demo videos provided by the Cityscapes dataset. The networks were trained in a leave-one-out scheme, where they were trained on two of the three videos and the test was performed on the third, unseen video. The training and testing was done for each combination of the three videos and the test results on the respective unseen video were averaged. 
For the exemplary test we limit ourself to the demo videos of the Cityscapes dataset, since our segmentation network was trained on this dataset and therefore we can definitely expect good performance on images from this domain.\\
For the \texttt{PredNet}, we used the standard architecture and training protocol as described by Lotter et al.\ \cite{Lotter2016}. For the predictive AE we adopted the architecture proposed by Hasan et al.\ \cite{Hasan2016} with some improvements to refine the predictions. The architecture is shown in Figure 4.
Typically a normal AE is trained using the MSE loss. It was shown that this leads to blurry predictions \cite{Mathieu2015,Lotter2016a}, which can be overcome by incorporating an adversarial loss. We added a discriminator network to the training procedure. The discriminator network uses the same architecture as the encoder network of the AE, extended with a patch-wise classification (also known as local adversarial loss) as proposed by \cite{Shrivastava2017}, where the discriminator network is trained to classify the real images or predicted images of the image prediction network w.r.t.\ the respective classes, with $s^{\mathrm{(D)}} \in \mathcal{S}^{\mathrm{(D)}} = \left\lbrace s_{\mathrm{real}},s_{\mathrm{predicted}} \right\rbrace$ being the class upon which the discriminator decides. The architecture of the discriminator network is shown in Fig. 5.
\begin{figure}[tp!]
	\centering
	\includegraphics{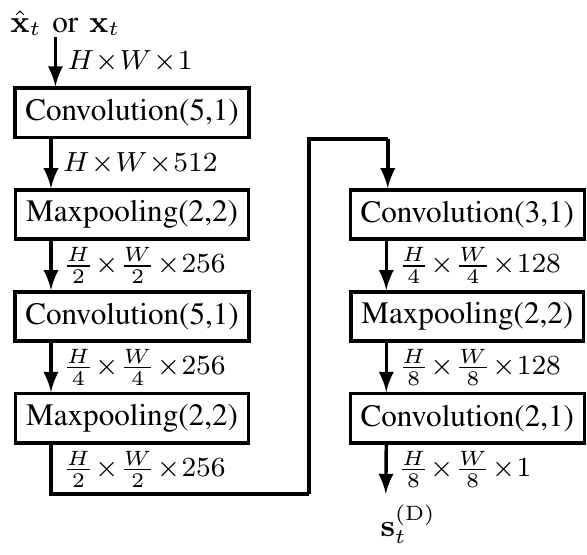}
	\caption{Architecture of the \textbf{discriminator} used for the adversarial loss of the \textbf{prediction training}. A convolution or maxpooling block with parameters ($K,S$) uses a filter size of $K \times K$ and a stride of $S$. The output class map $\mathbf{s}_t^{\mathrm{(D)}}$ of size $\frac{H}{8} \!\times\! \frac{W}{8}$ provides a class $s^{\mathrm{(D)}}$ for $8\!\times\!8$ pixel squares of the image.}
	\label{fig:short}
\end{figure}
\begin{figure*}[tp!]
	\centering
	\includegraphics{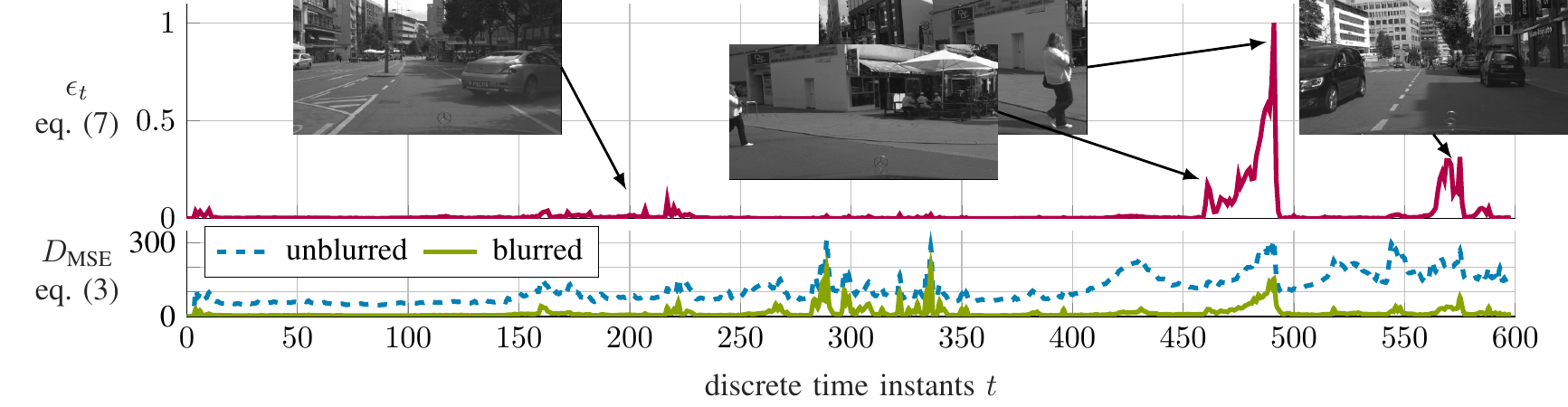}
	\caption{Exemplary results of MSE and the corner case score on the demo video \texttt{stuttgart\_00} of the Cityscapes dataset. The lower subfigure shows the MSE (3) for unblurred (blue, dashed) and blurred (green, solid) images. The upper subfigure shows the corner case score (7) that is based on the MSE computed with blurred images (red, solid). Marked are different situations yielding medium to high corner case scores and are therefore possible corner cases. 
	}
	\label{fig:short}
\end{figure*}
The loss of the generator and of the discriminator are added in the loss function
\begin{equation}
\label{eq:cost_func}
J_{\mathrm{G}}(\mathcal{A}_{\mathrm{G}},\mathcal{A}_{\mathrm{D}})=J_{\mathrm{MSE}}(\mathcal{A}_{\mathrm{G}})+\lambda J_{\mathrm{ADV}}(\mathcal{A}_{\mathrm{D}},\mathcal{A}_{\mathrm{G}}),
\end{equation}
with $J_{\mathrm{MSE}}$ being the standard MSE loss, $J_{\mathrm{ADV}}$ being the adversarial loss, and the weighting factor $\lambda = 0.25$ determining the influence of the adversarial loss on the overall loss. $\mathcal{A}_{\mathrm{G}}$ and $\mathcal{A}_{\mathrm{D}}$ denote the weights of the generator network, in this case the image prediction network, and the discriminator, respectively.\\
We found a major improvement in stability of the training by using a cyclic learning rate as proposed by \cite{Smith2017}. We used the stochastic gradient descent and started the training with a learning rate $\eta_{\mathrm{start, G}}=1\cdot10^{-7}$ of the generator and then periodically increased the learning rate of the generator to $\eta_{\mathrm{max, G}}=7\cdot10^{-7}$ and decreased it back again. The period $\phi = 20 \text{ epochs}$ showed the best results. The learning rate of the discriminator did not follow the cyclic protocol and was set to $\eta_{\mathrm{D}}=1\cdot10^{-6}$. The maximum learning rate is also subject to an exponential decay after each epoch with an exponent of $\gamma=0.85$. The batch size for the training is set to $b=3$. For the training, the original Cityscapes images were downsampled to a resolution of $H = 256 \text{ and } W = 512$ pixels and converted to greyscale.\\
The results in Table II show that the quality of the predictions from both networks are not far apart from each other for both PSNR and SSIM. In our experiment, however, the predictive autoencoder provides clearly a better MSE.
We decide to use the predictive adversarial AE, since it is a simple feed-forward architecture that can be easily modified for further experiments and also provides the better MSE, which is closely related to the measure that we use for corner case detection. \\
\subsection{Detection System}
As already mentioned, the task of corner case detection was an ill-defined problem lacking useful metrics so far. This is the reason why our approach to corner case detection relies on a clear definition of what a corner case is (Fig. 2), which is then conceptually implemented by a respective modular structure (Fig. 3), where two of the modules follow well-known quality metrics (see Sections IV.A and IV.B).\\
Typical metrics for the third module, namely the detection system, such as the false acceptance rate (FAR), false rejection rate (FRR) or the area under curve (AUC) of the receiver operation characteristic (ROC) can only be applied if (human-)labeled test data is available. To the best of our knowledge, there is no dataset of labeled corner cases for autonomous driving available so far. Therefore, in this work, we show exemplary results that were achieved on demo video material from the Cityscapes dataset.\\
Fig. 6 shows some exemplary results of $D_\text{MSE}(\hat{\mathbf{x}}_{t}, \mathbf{x}_{t})$~(3) (lower subfigure) and the final corner case score $\epsilon_t$~(7) (upper subfigure) on the demo video \texttt{stuttgart\_00} of the Cityscapes dataset. A problem with the image prediction module is that the squared error $e_t^2(i)$ was considerably higher in image regions with high frequencies. 
\begin{table}[tp!]
	\centering
	\caption{Example results for the \texttt{PredNet} and the predictive autoencoder on the three demo videos of the Cityscapes dataset. The results are averaged over the unseen test videos of a leave-one-out training on the Cityscapes dataset. Note that the number of parameters of both networks are quite close. During the training of the predictive autoencoder the discriminator parameters will increase the total amount of parameters to 10,795,650. Best values are printed in \textbf{bold}:}
	\resizebox{\columnwidth}{!}{%
		\begin{tabular}{ccccc}\toprule
			Model & MSE & PSNR& SSIM &\# Parameters\\
			\midrule
			\texttt{PredNet}	& 319.19 & 23.41  & \textbf{0.795}&6,909,818 \\[1ex]
			Pred. AE	& \textbf{209.78} & \textbf{25.53}  & 0.721&7,209,089 \\
			\bottomrule
		\end{tabular}
	}
	\label{tab:pred_results}	
\end{table}
We have therefore low-pass-filtered both the real image and the predicted image with a Gaussian kernel filter with kernel  size $10\!\times\!10$ and calculated the squared error afterwards. In the lower subfigure the MSE (3) is depicted without such blurring (blue, dashed) and with blurring (green, solid). The effect of blurring both the predicted image and the real image with a Gaussian kernel filter before subtraction (2) is that it helps to achieve  $D_\text{MSE} \approx 0$ in the many ordinary situations, while potential corner cases lead to fewer clearly observable periods of $D_\text{MSE}$ being larger than zero.\\
The plot in the upper subfigure (red, solid) shows the corner case score $\epsilon_t$ that is calculated according to (7). The corner case score is based on the MSE of the blurred images (6). It can be seen, that the system focuses on the relevant classes and suppresses high MSE values for non-relevant objects, when there are either no relevant classes or they are all predictable. Marked are three exemplary situations yielding medium to high corner case scores. In the first situation, the preceding car is turning right and the driver of the ego vehicle is taking a slight left turn to overtake the other car. In the second situation, a pedestrian is suddenly crossing the street in a road curve, a potentially dangerous situation. The third situation is an oncoming car that is not predictable for our system and close to the ego vehicle, which in our case leads to a high corner case score. It can be seen that the woman immediately crossing in a road curve produces the highest corner case score in our system, a situation, which indeed can be considered a corner case.   
\section{CONCLUSIONS}
In this work we proposed a formal definition for a corner case that is applicable to autonomous driving. We consider that a corner case is given, if there is a \textit{non-predictable} \textit{relevant object/class} in \textit{relevant location}. Each of the three aspects of our definition is covered by a module in the proposed corner case detector. The semantic segmentation to identify relevant objects and the image prediction are both subsystems that can be improved on their own. Their performance can be easily compared to knew methods due to widely accepted metrics in the respective research fields. For the third subsystem, the detection system, we presented a conceptual framework, which showed promising results in preliminary qualitative experiments. It also showed the urgent need for data that covers labeled corner cases. Accordingly, the next step of our work is to record videos of (arranged) corner cases along with labels in order to be able to also quantitatively evaluate the entire detection system. \\
\section{ACKNOWLEDGMENT}
The authors gratefully acknowledge support of this work by Volkswagen Group Research, Wolfsburg, Germany.

\addtolength{\textheight}{-14.8cm}   







{\small
	\bibliographystyle{IEEEbib}
	\bibliography{bolte_ownbib}

\begin{thebibliography}{10}

\bibitem{NTSB2018}
National Transportation~Safety Board,
\newblock ``{Preliminary Report Highway HWY18MH010},'' May 2018.

\bibitem{NTSB2018a}
National Transportation~Safety Board,
\newblock ``{Preliminary Report Highway HWY18FH011},'' June 2018.

\bibitem{Tian2018}
Y.~Tian, K.~Pei, S.~Jana, and B.~Ray,
\newblock ``{DeepTest: Automated Testing of Deep-Neural-Network-Driven
  Autonomous Cars},''
\newblock {\em arXiv preprint arXiv:1708.08559}, 2017.

\bibitem{Koopman2016}
P.~Koopman and M.~Wagner,
\newblock ``{Challenges in Autonomous Vehicle Testing and Validation},''
\newblock {\em SAE International Journal of Transportation Safety}, vol. 4, no.
  1, pp. 15--24, Apr. 2016.

\bibitem{Ulbrich2015}
S.~Ulbrich, T.~Menzel, A.~Reschka, F.~Schuldt, and M.~Maurer,
\newblock ``{Defining and Substantiating the Terms Scene, Situation, and
  Scenario for Automated Driving},''
\newblock in {\em Proc. of ITS}, Canary Islands, Spain, Sept. 2015, pp.
  982--988.

\bibitem{Chawla2002}
N.~V. Chawla, K.~W. Bowyer, L.~O. Hall, and W.~P. Kegelmeyer,
\newblock ``Smote: Synthetic minority over-sampling technique,''
\newblock {\em Journal of Artificial Intelligence Research (JAIR)}, vol. 16,
  pp. 321--357, 2002.

\bibitem{Chandola2009}
V.~Chandola, A.~Banerjee, and V.~Kumar,
\newblock ``{Anomaly Detection: A Survey},''
\newblock {\em ACM Computing Surveys}, vol. 41, no. 3, pp. 15:1--15:58, July
  2009.

\bibitem{Pimentel2014}
M.~Pimentel, D.~Clifton, L.~Clifton, and L.~Tarassenko,
\newblock ``{A Review of Novelty Detection},''
\newblock {\em Signal Processing}, vol. 99, pp. 215 -- 249, June 2014.

\bibitem{Hodge2004}
V.~Hodge and J.~Austin,
\newblock ``{A Survey of Outlier Detection Methodologies},''
\newblock {\em Artificial Intelligence Review}, vol. 22, no. 2, pp. 85--126,
  Oct 2004.

\bibitem{Thompson2002}
B.~B. Thompson, R.~J. Marks, J.~J. Choi, M.~A. El-Sharkawi, M.-Y. Huang, and
  C.~Bunje,
\newblock ``{Implicit Learning in Autoencoder Novelty Assessment},''
\newblock in {\em Proc. of IJCNN}, Honolulu, HI, USA, May 2002, vol.~3, pp.
  2878--2883 vol.3.

\bibitem{Singh2004}
S.~Singh and M.~Markou,
\newblock ``{An Approach to Novelty Detection Applied to the Classification of
  Image Regions},''
\newblock {\em IEEE Transactions on Knowledge and Data Engineering}, vol. 16,
  no. 4, pp. 396--407, April 2004.

\bibitem{Mahadevan2010}
V.~Mahadevan, W.~Li, V.~Bhalodia, and N.~Vasconcelos,
\newblock ``{Anomaly Detection in Crowded Scenes},''
\newblock in {\em Proc. of CVPR}, San Francisco, CA, USA, June 2010, pp.
  1975--1981.

\bibitem{Hasan2016}
M.~Hasan, J.~Choi, J.~Neumann, A.~K. Roy-Chowdhury, and L.~S. Davis,
\newblock ``{Learning Temporal Regularity in Video Sequences},''
\newblock in {\em Proc. of CVPR}, Las Vegas, NV, USA, June 2016, pp. 733--742.

\bibitem{Carrera2017}
D.~Carrera, F.~Manganini, G.~Boracchi, and E.~Lanzarone,
\newblock ``Defect detection in sem images of nanofibrous materials,''
\newblock {\em IEEE Transactions on Industrial Informatics}, vol. 13, no. 2,
  pp. 551--561, April 2017.

\bibitem{Chong2017}
Y.~S. Chong and Y.~H. Tay,
\newblock ``{Abnormal Event Detection in Videos Using Spatiotemporal
  Autoencoder},''
\newblock in {\em Proc. of 14th International Symposium on Neural Networks
  (ISNN)}, Hokkaido, Japan, June 2017, pp. 189--196.

\bibitem{Utkin2017}
L.~V. Utkin, V.~S. Zaborovsky, A.~A. Lukashin, S.~G. Popov, and A.~V.
  Podolskaja,
\newblock ``A siamese autoencoder preserving distances for anomaly detection in
  multi-robot systems,''
\newblock in {\em Proc. of ICCAIRO}, May 2017, pp. 39--44.

\bibitem{Munawar2017}
A.~Munawar, P.~Vinayavekhin, and G.~D. Magistris,
\newblock ``{Spatio-Temporal Anomaly Detection for Industrial Robots Through
  Prediction in Unsupervised Feature Space},''
\newblock in {\em Proc. of 2017 IEEE Winter Conference on Applications of
  Computer Vision (WACV)}, Santa Rosa, CA, USA, March 2017, pp. 1017--1025.

\bibitem{Liu2018}
W.~Liu, W.~Luo, D.~Lian, and S.~Gao,
\newblock ``{Future Frame Prediction for Anomaly Detection – A New
  Baseline},''
\newblock in {\em Proc. of CVPR}, Salt Lake City, UT, USA, June 2018, pp.
  6536--6545.

\bibitem{Lin2010}
R.~Lin, E.~Khalastchi, and G.~A. Kaminka,
\newblock ``{Detecting Anomalies in Unmanned Vehicles Using the Mahalanobis
  Distance},''
\newblock in {\em Proc. of ICRA}, Anchorage, AK, USA, May 2010, pp. 3038--3044.

\bibitem{Lampiri2017}
E.~Lampiri,
\newblock ``{Sensor Anomaly Detection and Recovery in a Nonlinear Autonomous
  Ground Vehicle Model},''
\newblock in {\em Proc. of ASCC}, Gold Coast, Australia, Dec 2017, pp.
  430--435.

\bibitem{Egner2010}
T.~Egner, J.~M. Monti, and C.~Summerfield,
\newblock ``{Expectation and Surprise Determine Neural Population Responses in
  the Ventral Visual Stream},''
\newblock {\em Journal of Neuroscience}, vol. 30, no. 49, pp. 16601--16608,
  2010.

\bibitem{Ranzato2014}
M.~A. Ranzato, A.~Szlam, J.~Bruna, M.~Mathieu, R.~Collobert, and S.~Chopra,
\newblock ``{Video (Language) Modeling: A Baseline for Generative Models of
  Natural Videos},''
\newblock {\em arXiv preprint arXiv:1412.6604}, 2014.

\bibitem{Srivastava2015}
N.~Srivastava, E.~Mansimov, and R.~Salakhudinov,
\newblock ``{Unsupervised Learning of Video Representations Using LSTMs},''
\newblock in {\em Proc. of ICML}, Lille, France, July 2015, pp. 843--852.

\bibitem{Mathieu2015}
M.~Mathieu, C.~Couprie, and Y.~LeCun,
\newblock ``{Deep Multi-Scale Video Prediction Beyond Mean Square Error},''
\newblock {\em arXiv preprint arXiv:1511.05440}, 2015.

\bibitem{Patraucean2015}
V.~Patraucean, A.~Handa, and R.~Cipolla,
\newblock ``{Spatio-Temporal Video Autoencoder With Differentiable Memory},''
\newblock {\em arXiv preprint arXiv:1511.06309}, 2015.

\bibitem{Lu2017}
C.~Lu, M.~Hirsch, and B.~Sch{\"o}lkopf,
\newblock ``{Flexible Spatio-Temporal Networks for Video Prediction},''
\newblock in {\em Proc. of CVPR}, Honolulu, HI, USA, July 2017, pp. 2137--2145.

\bibitem{Lotter2016}
W.~Lotter, G.~Kreiman, and D.~Cox,
\newblock ``{Deep Predictive Coding Networks for Video Prediction and
  Unsupervised Learning},''
\newblock {\em arXiv preprint arXiv:1605.08104}, 2016.

\bibitem{Long2015}
J.~Long, E.~Shelhamer, and T.~Darrell,
\newblock ``{Fully Convolutional Networks for Semantic Segmentation},''
\newblock in {\em Proc. of CVPR}, Boston, MA, USA, June 2015, pp. 3431--3440.

\bibitem{Russakovsky2015}
O.~Russakovsky, J.~Deng, H.~Su, J.~Krause, S.~Satheesh, S.~Ma, Z.~Huang,
  A.~Karpathy, A.~Khosla, M.~Bernstein, A.~C. Berg, and L.~Fei-Fei,
\newblock ``{ImageNet Large Scale Visual Recognition Challenge},''
\newblock {\em International Journal of Computer Vision (IJCV)}, vol. 115, no.
  3, pp. 211--252, 2015.

\bibitem{Zhao2016a}
H.~Zhao, J.~Shi, X.~Qi, X.~Wang, and J.~Jia,
\newblock ``{Pyramid Scene Parsing Network},''
\newblock in {\em Proc. of CVPR}, Honulu, HI, USA, July 2017, pp. 2881--2890.

\bibitem{Chen2017}
L.-C. Chen, G.~Papandreou, F.~Schroff, and H.~Adam,
\newblock ``{Rethinking Atrous Convolution for Semantic Image Segmentation},''
\newblock {\em {arXiv preprint arXiv:1706.05587}}, 2017.

\bibitem{Wu2016}
Z.~Wu, C.~Shen, and A.~van~den Hengel,
\newblock ``{Wider or Deeper: Revisiting the ResNet Model for Visual
  Recognition},''
\newblock {\em arXiv preprint arXiv:1611.10080}, 2016.

\bibitem{Chen2018a}
L.-C. Chen, G.~Papandreou, I.~Kokkinos, K.~Murphy, and A.~L. Yuille,
\newblock ``{Deeplab: Semantic Image Segmentation With Deep Convolutional Nets,
  Atrous Convolution, and Fully Connected CRFs},''
\newblock {\em IEEE Transactions on Pattern Analysis and Machine Intelligence},
  vol. 40, no. 4, pp. 834--848, Apr. 2018.

\bibitem{RotaBulo2018}
S.~R. Bul{\`o}, L.~Porzi, and P.~Kontschieder,
\newblock ``{In-Place Activated BatchNorm for Memory-Optimized Training of
  DNNs},''
\newblock in {\em Proc. of CVPR}, Salt Lake City, UT, USA, Jun 2018, pp.
  5639--5647.

\bibitem{He2016}
K.~He, X.~Zhang, S.~Ren, and J.~Sun,
\newblock ``{Deep Residual Learning for Image Recognition},''
\newblock in {\em Proc. of CVPR}, Las Vegas, NV, USA, July 2016, pp. 770--778.

\bibitem{Yu2016}
F.~Yu and V.~Koltun,
\newblock ``{Multi-Scale Context Aggregation by Dilated Convolutions},''
\newblock in {\em Proc. of ICLR}, San Juan, PR, May 2016.

\bibitem{Chen2018}
L.~C. Chen, G.~Papandreou, I.~Kokkinos, K.~Murphy, and A.~L. Yuille,
\newblock ``{DeepLab: Semantic Image Segmentation with Deep Convolutional Nets,
  Atrous Convolution, and Fully Connected CRFs},''
\newblock {\em IEEE Transactions on Pattern Analysis and Machine Intelligence},
  vol. 40, no. 4, pp. 834--848, April 2018.

\bibitem{Romera2018}
E.~Romera, J.~M. Álvarez, L.~M. Bergasa, and R.~Arroyo,
\newblock ``{ERFNet: Efficient Residual Factorized ConvNet for Real-Time
  Semantic Segmentation},''
\newblock {\em IEEE Transactions on Intelligent Transportation Systems}, vol.
  19, no. 1, pp. 263--272, Jan 2018.

\bibitem{Ronneberger2015}
O.~Ronneberger, P.~Fischer, and T.~Brox,
\newblock ``{U-Net: Convolutional Networks for Biomedical Image
  Segmentation},''
\newblock in {\em Proc. of MICCAI}, Munich, Germany, October 2015, pp.
  234--241, Springer International Publishing.

\bibitem{Ioffe2015}
S.~Ioffe and C.~Szegedy,
\newblock ``{Batch Normalization: Accelerating Deep Network Training by
  Reducing Internal Covariate Shift},''
\newblock in {\em Proc. of ICML}, Lille, France, July 2015, pp. 448--456.

\bibitem{Sandler2018}
M.~Sandler, A.~G. Howard, M.~Zhu, A.~Zhmoginov, and L.{-}C. Chen,
\newblock ``{MobileNetV2{:} Inverted Residuals and Linear Bottlenecks},''
\newblock in {\em Proc. of CVPR}, Salt Lake City, UT, USA, June 2018, pp.
  4510--4520.

\bibitem{Everingham2015}
M.~Everingham, S.~M. Eslami, L.~Gool, C.~K. Williams, J.~Winn, and
  A.~Zisserman,
\newblock ``{The Pascal Visual Object Classes Challenge: A Retrospective},''
\newblock {\em International Journal of Computer Vision}, vol. 111, no. 1, pp.
  98--136, Jan. 2015.

\bibitem{Gao2017}
H.~Gao, H.~Yuan, Z.~Wang, and S.~Ji,
\newblock ``{Pixel Deconvolutional Networks},''
\newblock {\em arXiv preprint arXiv:1705.06820}, 2017.

\bibitem{Wang2004}
Z.~Wang, A.~C. Bovik, H.~R. Sheikh, and E.~P. Simoncelli,
\newblock ``{Image Quality Assessment: From Error Visibility to Structural
  Similarity},''
\newblock {\em IEEE Transactions on Image Processing}, vol. 13, no. 4, pp.
  600--612, April 2004.

\bibitem{Zhang2015}
R.~Zhang, S.~A. Candra, K.~Vetter, and A.~Zakhor,
\newblock ``{Sensor Fusion for Semantic Segmentation of Urban Scenes},''
\newblock in {\em Proc. of ICRA}, Seattle, WA, USA, May 2015, pp. 1850--1857.

\bibitem{Wei2018}
P.~Wei, L.~Cagle, T.~Reza, J.~Ball, and J.~Gafford,
\newblock ``{LiDAR and Camera Detection Fusion in a Real-Time Industrial
  Multi-Sensor Collision Avoidance System},''
\newblock {\em Electronics}, vol. 7, no. 6, 2018.

\bibitem{Bosnak2017}
M.~Bosnak and I.~Skrjanc,
\newblock ``{Efficient Time-To-Collision Estimation for a Braking Supervision
  System with LIDAR},''
\newblock in {\em Proc. of CYBCONF}, Exeter, United Kingdom, June 2017, pp.
  1--6.

\bibitem{Cordts2016}
M.~Cordts, M.~Omran, S.~Ramos, T.~Rehfeld, M.~\mbox{Enzweiler}, R.~Benenson,
  U.~Franke, S.~Roth, and B.~Schiele,
\newblock ``{The Cityscapes Dataset for Semantic Urban Scene Understanding},''
\newblock in {\em Proc. of CVPR}, Las Vegas, NV, USA, June 2016, pp.
  3213--3223.

\bibitem{Lotter2016a}
W.~Lotter, G.~Kreiman, and D.~Cox,
\newblock ``{Unsupervised Learning of Visual Structure using Predictive
  Generative Networks},''
\newblock {\em arXiv preprint arXiv:1511.06380}, 2015.

\bibitem{Shrivastava2017}
A.~Shrivastava, T.~Pfister, O.~Tuzel, J.~Susskind, W.~Wang, and R.~Webb,
\newblock ``{Learning from Simulated and Unsupervised Images through
  Adversarial Training},''
\newblock in {\em Proc. of CVPR}, Honolulu, HI, USA, July 2017, pp. 2242--2251.

\bibitem{Smith2017}
L.~N. Smith,
\newblock ``{Cyclical Learning Rates for Training Neural Networks},''
\newblock in {\em Proc. of WACV}, Santa Rosa, CA, USA, March 2017, pp.
  464--472.

\end{thebibliography}
}

\end{document}